\documentclass[runningheads]{llncs}
\usepackage[T1]{fontenc}
\usepackage{graphicx}
\usepackage{comment}
\begin{document}
\title{A Comparative Study of Rule-Based and Data-Driven Approaches in Industrial Monitoring}
\titlerunning{Rule-Based and Data-Driven Approaches in Industrial Monitoring}
%
\author{Giovanni De Gasperis\inst{1}\orcidID{0000-0001-9521-4711} \and
Sante Dino Facchini\inst{1}\orcidID{0000-0002-2009-5209}
}
\authorrunning{G. De Gasperis et al.}
%
\institute{Università degli Studi dell'Aquila, Via Vetoio snc, IT671000 L'Aquila, Italy 
\email{giovanni.degasperis@univaq.it} \email{santedino.facchini@graduate.univaq.it}\\
}
\maketitle              
\begin{abstract}
Industrial monitoring systems, especially when deployed in Industry 4.0 environments, are experiencing a shift in paradigm from traditional rule-based architectures to data-driven approaches leveraging machine learning and artificial intelligence. This study presents a comparison between these two methodologies, analyzing their respective strengths, limitations, and application scenarios, and proposes a basic framework to evaluate their key properties. Rule-based systems offer high interpretability, deterministic behavior, and ease of implementation in stable environments, making them ideal for regulated industries and safety-critical applications. However, they face challenges with scalability, adaptability, and performance in complex or evolving contexts. Conversely, data-driven systems excel in detecting hidden anomalies, enabling predictive maintenance and dynamic adaptation to new conditions. Despite their high accuracy, these models face challenges related to data availability, explainability, and integration complexity. The paper suggests hybrid solutions as a possible promising direction, combining the transparency of rule-based logic with the analytical power of machine learning. Our hypothesis is that the future of industrial monitoring lies in intelligent, synergic systems that leverage both expert knowledge and data-driven insights. This dual approach enhances resilience, operational efficiency, and trust, paving the way for smarter and more flexible industrial environments.
\end{abstract}
\section{Introduction}\label{sec1}
In the era of Industry 4.0, characterized by the spread of sensors, Internet of Things (IoT) and Big Data, the industrial monitoring of machinery and processes is experiencing a profound transformation. Traditional Rule-based systems, which operate according to logics determined by human experts, are increasingly flanked by Data-driven approaches that employ Machine Learning (ML), Deep Learning (DL) and advanced statistical models \cite{rosati2023}. Numerous reviews highlight how the integration of Artificial Intelligence into industrial processes can significantly improve efficiency, quality and the ability to identify anomalies in real time. For example, a recent review of Leite et al. \cite{leite2024fault}, points out that the adoption of ML algorithms has already brought measurable benefits in terms of reducing unplanned failures and optimizing maintenance. At the same time, the literature reports that companies continue to make extensive use of systems based on knowledge and previous experience, especially when high levels of reliability and explainability are required. 
In this report, the two approaches rule-based monitoring systems vs data-driven systems are compared in detail, analysing their advantages, disadvantages, performance and application scenarios in the industrial field, with references to the most relevant scientific literature.

\section{Rule Based Approach to Monitoring Systems}\label{sec2}
Rule-based monitoring systems use explicit domain knowledge that is encoded in the form of logical rules, thresholds, or predefined decision trees. In practice, a human expert (e.g. a maintenance or process engineer) working with a knowledge engineer can define a set of if-then rules that describe normal or fault conditions \cite{bernard2002}. For example, you can establish that "IF the temperature exceeds 80°C FOR more than 10 minutes, THEN generate an alarm". These rules can be derived from physical laws, from the manufacturer's manuals or from empirical experience accumulated over time. Systems of this type fall into the category of expert systems and can also include fuzzy logics (stepwise values) to handle uncertainty and tolerances. A classic example is the fuzzy expert system for diagnosing faults in electrical systems described in Monsef et al. \cite{monsef1997fuzzy}, in which fuzzy rules codified by electrical engineers detect faults on the power grid.

\subsection{Benefits of Rule-based Systems}\label{subsec2}
We will start our analysis considering the benefits that Rule-based system can apport to industrial monitoring systems.
\subsubsection{Interpretability and transparency.} Every decision or alarm generated can be explained by the corresponding rule. This makes the system easy to understand for operators and technicians. The rules make human knowledge explicit, ensuring that the behavior of the monitoring system is transparent. In critical industrial settings, explainability is critical: for example, rule-based systems follow well-known procedures and industry regulations, aligning with established practices. The review of Bouhadra et al. \cite{bouhadra2024knowledge} highlights that knowledge-based methods formalize the know-how of experts and company standards, providing conclusions consistent with experience and facilitating user trust in the system .
\subsubsection{Ease of implementation (in known contexts).} If the process to be monitored is well understood and the expected anomalous conditions are few and well defined, then writing rules is relatively simple. Many traditional Supervisory Control and Data Acquisition (SCADA) systems and factory Programmable Logic Controllers (PLCs) use threshold alarms for this very reason. Boolean logic is intuitive for operators: in the Heating, Ventilation, and Air Conditioning (HVAC) industry, for example, Boolean rules (e.g., temperature thresholds) are the current standard in building Fault Detection \& Diagnostics (FDD) systems \cite{kim2018}. Research of Nelson et al. \cite{nelson2024machine} confirms that plant operators find it more immediate to interact with simple rules rather than complex statistical models, which is why today's commercial solutions often employ fixed rules.
\subsubsection{Reliability under expected conditions.} A well-designed rules-based system, validated by experts, will operate deterministically in the situations for which it was conceived. This is important in industries where safety is critical: rules ensure that certain limits are never exceeded without warning. For example, a chemical plant can implement interlocks and alarms based on rigid logic to prevent hazardous conditions. As long as the plant operates within the perimeter of the envisaged scenarios, the rules ensure rapid and safe responses, often with minimal computational requirements (they can run on dedicated control units with limited power, e.g. PLC) \cite{musulin2013}.
\subsubsection{Low dependency on historical data.} In contrast to Data-driven methods, Rule-based systems do not require large datasets of failure examples to develop. They are based on engineering knowledge already available; this makes them suitable in contexts where failures are very rare (e.g. aerospace) and there is not enough data to train a statistical model. In these cases, it is more practical to rely on expert insights and simulations to enumerate possible anomalous conditions \cite{islam2018}.

\subsection{Disadvantages of rule-based systems}\label{subsubsec2}
Coming to the negative aspects of Rule-based systems, we can refer to the following issues.

\subsubsection{Limited coverage and rigidity.} For each decision process or alarm generated, this approach requires a corresponding rule that explicitly defines the system’s behavior in that specific situation. This makes the system transparent and easy to interpret for operators and technicians. However, it also makes the system difficult to adapt when changes occur in the control process or when new sensors need to be integrated. While the rules capture and formalize human knowledge, they are often highly domain-specific. In critical industrial settings, where the process may not be fully understood, this limitation can lead to a rapid decline in the effectiveness of fault detection and parameter monitoring. 

\subsubsection{Scalability and difficult maintenance.} As system complexity and the number of monitored variables increase, the required number of rules required grows rapidly. Managing hundreds of interdependent rules becomes onerous; Updating the system requires manual intervention by an expert to add/modify rules, with the risk of introducing errors or inconsistencies. Design requires in-depth knowledge: it is necessary to extract and formalize the tacit knowledge of engineers, a non-trivial process (the so-called bottleneck knowledge acquisition). In addition, if the process changes (e.g. a new machine or product is introduced), many rules often need to be revised. In the literature, cases are reported in which expert systems become difficult to maintain in the long term precisely because of the accumulation of rules and exceptions. The study by Nelson et al. on FDDs \cite{nelson2024machine} in the building field highlights that the rules must be defined and calibrated individually by experts for each scenario, and each rule covers only that specific behaviour beyond threshold: this indicates low flexibility and adaptability to new or slightly different situations.

\subsubsection{Suboptimal performance in complex scenarios.} When the phenomena to be monitored depend on many variables with non-linear relationships, a simple set of rules can fail to capture the complexity of the system. For example, in the monitoring of an advanced production line, an anomaly can manifest as a slight multivariate pattern deviation (correlated change in temperature, vibration, and pressure simultaneously) a situation that is difficult to capture with an isolated if-then rule for each signal. Rule-based systems, typically focused on single parameters or discrete conditions, can generate both false negatives (missed detections) in the presence of complex anomalous patterns, and false positives if thresholds are set too conservatively. Fine calibration of thresholds takes time and often remains a trade-off between sensitivity and specificity.
\\
\begin{table}[h!]
\caption{Advantages, Disadvantages, and Ideal Use Cases of Rule-based Monitoring systems.}\label{tab1}
\centering
\begin{tabular}{|p{2,5cm}|p{4,25cm}|p{4,25cm}|}
\hline
\textbf{Feature} & \textbf{Advantages} & \textbf{Disadvantages} \\
\hline
Data dependency & Work well even with
limited data, no need for training datasets & Do not automatically adapt to new data or contexts, need human a-priori knowledge \\
\hline
Transparency & Explicit and easily understandable rules & None \\
\hline
Maintainability \& Adaptability & Easy to modify in simple and known environments & Difficult to manage and maintain in systems with many rules (accumulation problem) \\
\hline
Scalability & Good in restricted and well-defined domains & Limited in dynamic or complex environments \\
\hline
Performance & High in deterministic tasks & Not optimal in probabilistic or ambiguous tasks \\
\hline
Generalization & Poor, each rule is task specific & Unable to infer new knowledge from the data \\
\hline
\textbf{Ideal Use Cases} & \multicolumn{2}{p{8,5cm}|}{
- Known systems where knowledge is full and consolidated \newline
- Diagnostics and early warning in well-defined procedures \newline
- Contexts with stable rules and few changes \newline
- Applications where transparency is crucial (e.g., industrial processes where auditing and compliance required by law)
} \\
\hline
\end{tabular}
\end{table}

In summary, rule-based approaches work well in stationary and well understood domains, where failure modes are already known, and the primary goal is transparency and ease of validation. However, they show their limits in highly variable or innovative contexts, in which being able to dynamically discover new forms of anomaly is crucial. In Table \ref{tab1} we present a quick reference table that summarize positive and negative aspects of Rule-based approach to industrial monitoring.

\section{Data-Driven Approaches to Monitoring Systems}\label{sec3}
Data-driven approaches to monitoring systems rely on the automatic analysis of historical and current data of an industrial process to identify patterns associated with normal or fault conditions. Instead of manually written rules, these methods learn from observations: they use ML algorithms, statistical techniques or DL to model plant behaviour. In practice, the system is provided with a set of training data (e.g. time recordings of sensors during normal operations and during known failures) that are used by the algorithm to build a mathematical model. The model is capable to distinguish the two conditions of the monitored system (normal or fault) and autonomously identifies when the data "does not resemble" typical data (unsupervised approach).
Let's examine deeper these two methods.

\subsubsection{Classical machine learning methods.} Consider classical techniques like Bayesian Networks, Decision Trees, Random Forests, Support Vector Machines (SVMs), k-nearest Neighbours, Clustering Models, etc. These algorithms can be both supervised (requiring data labelled as "faulty" vs "normal" to learn a classifier) and unsupervised (identify anomalies such as statistical deviations from normal patterns). Techniques such as Principal Component Analysis (PCA) or Feature Selection are part of the data-driven toolkit to reduce dimensionality and isolate relevant signals. A broad overview of such methodologies is provided by Ji et al. \cite{ji2022review}, which discusses both traditional statistical approaches and newer machine learning algorithms for industrial monitoring.

\subsubsection{Deep Learning Methods.} Deep Neural Networks (DNNs) have revolutionized the field in recent years. Architectures such as Convolutional Neural Networks (CNNs), Recurrent Neural Networks (RNNs) and their enhanced version Long Short-Term Memory (LSTMs) can automatically extract features from raw data (e.g., detect complex temporal patterns in signals, or visual features in production images) and build highly accurate predictive models. For example, in predictive maintenance, CNNs and LSTMs can analyse vibration or machine audio time series to detect incipient anomalies without the need to manually predefine condition indicators \cite{yang2020}. This end-to-end approach is considered a huge step forward from classical methods that required manual feature extraction by experts. The review of Vashishtha et al. \cite{vashishtha2025roadmap} underlines how the advent of deep learning has reformed Failure diagnostics processes, moving from a paradigm in which effectiveness depended on feature engineering by experts, to one in which the model learns directly from raw data, reducing human effort and fully automating the identification of machinery health. In other words, deep learning models are able to bridge the gap between the large amount of monitoring data collected (thanks to IoT sensors in the factory) and useful information on the condition of the plant, discovering hidden correlations not evident to humans.

\subsection{Benefits of Data-driven Approaches}\label{subsec_advDDA}
We can now examine the strengths of approaching Industrial monitoring following a Data-driven philosophy.

\subsubsection{Ability to detect complex patterns and unexpected failures.} 
One of the greatest strengths of data-driven methods is their ability to discover correlations and weak signals in data that would escape a manual approach. ML algorithms can combine tens or hundreds of process variables at once, identifying multivariate anomalies. This makes it possible to intercept failures at a very early stage, when the symptoms on individual variables are still mild but the combination with other signals reveals an abnormal trend \cite{Jieyang2023}. In addition, these systems can also recognize new faults, never seen before, especially if designed with unsupervised approaches (e.g., anomaly detection models that report any significant deviation from usual behavior). The literature confirms this capability: the review of Leite et Al. \cite{leite2024fault} mentioned above shows that data-driven techniques, by learning from operating data, can identify both known failure patterns and anomalous conditions never encountered before, thus extending the adaptability of monitoring beyond the initial knowledge codified by experts. Advanced algorithms are able to "generalize" and discover similarities with known failures even in slightly different scenarios, or to isolate totally new outliers. This reduces reliance on predefined rules and allows systems to adapt to a wider range of operating conditions.

\subsubsection{Better performance (accuracy, timeliness).} 
In many practical cases, ML/DL-based systems have demonstrated superior performance compared to fixed-threshold systems, both in terms of accuracy in fault recognition and in terms of reducing false alarms.  State-of-the-art DL approaches can outperform conventional detection systems, achieving significantly higher accuracy rates. For example, a paper of Shiney \cite{shiney} reports that a deep neural network model for detecting defects in industrial gaskets achieves significantly better results than traditional rule-based methods, dramatically reducing both missed detection errors and false alarms.  Also in the maintenance field, the integration of boosting algorithms (e.g. XGBoost) has shown to be able to significantly increase the accuracy in diagnosing faults compared to manual analysis with Boolean rules \cite{tang2020}. In addition, these models can often pick up weak signals and diagnose them before they become full-blown failures, improving the timeliness of maintenance interventions.

\subsubsection{Reduce manual work and discover knowledge.} 
A good data-driven model automates the analysis step that would otherwise be done manually by an expert by examining graphs and sensor logs. This reduces the human workload and potential human error in interpreting the data. As pointed out in the review of Vashishtha et Al. \cite{vashishtha2025roadmap}, advanced ML methods drastically reduce the human effort in identifying the health conditions of machines, realizing end-to-end automatic monitoring. In addition, such models can bring out new knowledge about the factors that contribute to failures: for example, an algorithm can find that certain current fluctuations, thought to be negligible, systematically precede a certain type of anomaly–information that can then be reverse-engineered by experts to improve understanding of the process. In this sense, data-driven systems not only monitor, but contribute to enriching the knowledge available on the system through data analysis.

\subsubsection{Flexibility and broad applicability.}
Once the data collection infrastructure is set up, the same algorithms can often be reused in different contexts with minimal changes. For example, an anomaly detection model based on autoencoder neural networks can be applied to both the monitoring of an electric motor and that of a pump, simply by changing the training dataset. This offers a certain generality: the review of Ige at al. \cite{ige2025machine} shows numerous successful cases of ML algorithms adopted in various sectors (manufacturing, energy, logistics) with similar benefits in terms of optimization and reduction of defects. This suggests that data-driven techniques constitute a sort of universal toolbox adaptable to multiple industrial problems, as opposed to rules, often sewn on a specific plant.

\subsection{Disadvantages of Data-driven Approaches}\label{subsec_disDDA}
On the other side there are several flaws in approaching industrial monitoring with Data-driven solutions.

\subsubsection{Poor interpretability (black box).}
A major problem with complex models (especially DNNs) is the difficulty of understanding how they make decisions. Unlike simple rules, ML models rely on hundreds of internal mathematical parameters that are not immediately understandable. This leads to a lack of transparency that can hinder the trust of operators and managers in the monitoring system. The survey of Bouhadra et Al.\cite{bouhadra2024knowledge} points out that the lack of explainability is one of the main limitations of data-driven methods in maintenance, precisely because it makes the analysis of the causes of identified faults complex. In areas such as aviation or medical, regulations often require a clear justification for each alarm to avoid false alarms that can lead to unnecessary downtime or invasive interventions, and black-box models struggle to meet these requirements. To overcome this, sub-disciplines such as Explainable AI (XAI) are emerging, which develop techniques to extract understandable explanations from ML models, but these solutions are still maturing \cite{gilpin2018}.

\subsubsection{Data dependency (quantity and quality).}
Data-driven systems require abundant and representative historical data to function well. Supervised methods require datasets with sufficient failure examples to train a robust classifier – which can be problematic, as relevant failures are rare events. Even for unsupervised or semi-supervised methods, it is necessary to have data that extensively covers normal behaviour and operational variability. Poor or incomplete data can lead to the model missing real failures or generating too many false alarms \cite{merim2021}. In addition, data quality is crucial: outliers in sensors, calibration errors, missing or noisy data can mislead machine learning. As highlighted by Saeed et al. \cite{saeed2025deep}, many academic methods assume that data are available complete, balanced and abundant, an assumption that is often unrealistic in real industrial scenarios. In practice, coverage of certain failure modes may be lacking, or the available data may be unbalanced (very few examples of failure vs. many of normal operation), making it difficult to train reliable models. Another review points out that the scarcity of data is a bottleneck in some sectors: for example, in plants where installing additional sensors is expensive or impractical, the information base remains limited. 

\subsubsection{Need for specialized skills.}
Developing and fine-tuning data-driven models requires data science and ML skills that are not always present in the traditional maintenance or automation team. A multidisciplinary approach is often required, involving data scientists together with domain engineers, to prepare the data, choose/optimize the algorithms and interpret the results. According to the review on FDD in Industry 4.0 cited before \cite{leite2024fault}, the limited experience in ML within industrial organizations represents a significant challenge to the large-scale implementation of these techniques. In addition, feature engineering remains partly necessary for classical methods: identifying which signals and indicators to provide to the model is a crucial task and requires technical knowledge of the process and statistics. For end-to-end DL methods, this burden is less, but the need to design the network architecture, training parameters, etc., remains tasks that require specific expertise. In summary, the initial cost in terms of specialized human resources can be high.
\subsubsection{Computational and implementation complexity. }
A sophisticated ML model can require significant computational resources, both for training (which is often performed on servers or clouds with GPUs) and, sometimes, for real-time inference if the model is very complex. In edge industrial contexts (e.g. embedded devices near machinery), implementing a deep neural network in real-time can pose latency and power consumption challenges \cite{bian2022}. The survey on deep learning and fault diagnosis  published in Scientific Reports \cite{shiney} addresses the issue of edge-cloud architectures to manage algorithms in environments with limited resources: the need to balance model accuracy with computational efficiency is an important practical aspect. In some cases, simpler or compressed models (model compression, quantization, distillation) are opted for deployment in the field, accepting a slight drop in performance to be able to run them on limited hardware. Also, from a software point of view, integrating a data-driven model into the control system requires interfacing with factory databases, data acquisition systems, and providing model update mechanisms: integration can therefore be more complex than a few hard-coded rules in a PLC.
\subsubsection{Risk of overfitting and limited adaptability outside of training conditions:}
Data-driven models learn from historical data: if the future operating environment differs significantly from the past (e.g., new products, process changes), the model may not generalize well and lose accuracy. A classic problem is the concept of drift: the distribution of data changes over time (e.g. due to a progressive deterioration of the plant), and the model trained on old data no longer recognizes the situation correctly. Rule-based systems, although rigid, can be more predictable outside the expected conditions (if a variable goes out of range unexpectedly, at least a generic alarm is triggered), while an ML model could give erratic outputs if you leave the training domain. To mitigate this, continuous monitoring of the model's performance and periodic re-training of the model with new data is required which closes the loop and brings us back to the need for up-to-date data and expertise to carry out this model maintenance. Without this, a data-driven system can degrade over time.

\begin{table}[h!]
\centering
\caption{Advantages, Disadvantages, and Ideal Use Cases of Data-driven Monitoring systems}
\label{tab2}
\begin{tabular}{|p{3cm}|p{4cm}|p{4cm}|}
\hline
\textbf{Feature} & \textbf{Advantages} & \textbf{Disadvantages} \\
\hline
Data Dependency & Can learn complex patterns from large datasets & Require large amounts of high-quality data \\
\hline
Transparency & Often considered black-box models & Hard to explain decisions to non-experts personnel, Lack transparency in decision-making processes \\
\hline
Maintainability \& Adaptability & Automatically adapt to new data and environments & May overfit or underperform with limited data \\
\hline
Scalability & Highly scalable with increasing data and computing power & Computationally intensive and resource-demanding \\
\hline
Performance & Excellent in probabilistic and ambiguous tasks & Difficult to interpret and debug \\
\hline
Generalization & Can generalize from examples to unseen cases & None \\
\hline
\textbf{Ideal Use Cases} & \multicolumn{2}{p{8cm}|}{
- Systems leveraging images and speech recognition \newline
- Predictive analytics and forecasting of faults \newline
- Large scale-monitoring \newline
- Autonomous systems and robotics \newline
- Hidden patterns discovery
} \\
\hline
\end{tabular}
\end{table}

In summary, data-driven methods represent the state of the art for addressing complex and large-scale monitoring problems, thanks to their effectiveness in capturing hidden patterns and adapting to data. However, they pose significant practical challenges around data, expertise, and trust, which can limit their adoption or effectiveness if not managed properly. In Table \ref{tab2} we present a quick reference table that summarize positive and negative aspects of Data-driven approach to industrial monitoring.

\section{An Evaluation Framework for Direct Comparison Between Rule-Based, Data-Driven and Hybrid Approaches}\label{sec4}
The idea of the paper is to propose also an evaluation framework to determine if a monitoring system based either in pure Data-driven or pure Rule-based or Hybrid has suitable characteristic for the adopting organization. We determined six parameters on the base of the previous analysis that are presented below.

\subsubsection{Interpretability and Transparency.} Rule-based systems excel in explainability as each alarm is justified by a readable condition. This is crucial in regulated environments (e.g. pharmaceutical or aerospace) where it is necessary to immediately understand why an alarm has been triggered. In contrast, data-driven approaches (especially DL) often act like black boxes, making it difficult for an engineer to understand which signals contributed to the decision. This can create mistrust and hinder adoption in areas where the stakes are high. As a result, in applications where justifiability of monitoring is essential (e.g. decision support systems for power plant operators), rule-based methods retain an advantage. Data-driven methods typically require the addition of interpretation modules (e.g. XAI techniques like feature attribution, local explanations) to fill this gap.
    
\subsubsection{Demand for data and a priori knowledge.} A rule-based system relies on human knowledge embedded ex-ante; it does not need to "see" large amounts of historical data to function, but it does require prior expert know-how. By contrast, a Data-driven system typically requires a lot of data to train or configure. This means that, to implement ML, the company must have collected quality data over time (or run test campaigns to generate it). In cases where such data does not exist, the barrier to entry for data-driven methods is high, and a rules-based approach may be the only feasible option initially. On the other hand, if the plant is new or innovative and there are no experts with direct experience, the rules cannot be written – whereas a data-driven method could learn directly from operational experience as data is collected. In practice, knowledge vs data represents a trade-off: rule-based systems transfer human knowledge to the system, data-driven systems extract knowledge from data.
    
\subsubsection{Flexibility and adaptation.} A rule set is inherently static and covers limited scenarios; if the process changes or new types of defects emerge, you must intervene manually to update the rules. On the contrary, a Data-driven model can be made adaptive: for example, through periodic retraining on the most recent data, it can follow the evolution of the monitored system. In addition, ML algorithms can theoretically span a wider range of situations than hard-coded rules. This makes data-driven approaches more suitable for contexts where variability or continuous innovation is expected (eg. multi-product production lines, mass customization, etc.). On the downside, this flexibility means that the data-driven model must be revalidated frequently, while the rules, once validated in a stable scenario, do not require change if the scenario remains so.

\subsubsection{Performance and accuracy of detection.} Numerous studies indicate that Data-driven methods  generally achieve better performance indicators than Rule-based systems with adequate data. This means higher sensitivity (ability to catch even difficult faults) and specificity (lower false alarm rate). For example, in a comparison of HVAC systems, ML algorithms such as decision trees identified operating anomalies that had not been covered by standard Boolean rules, showing a distinct advantage in diagnosing nuanced problems ref(35). A further advantage is in early detection: predictive models can anticipate incipient failures (e.g. by estimating the remaining useful life of a component) while a rule-based system typically reports only the full-blown failure when a threshold is exceeded. However, there are cases in which well-calibrated rule systems have comparable or even better performance on specific simple problems: if the phenomenon to be detected is one-dimensional and very well defined, a simple rule can be optimized to have zero false negatives and very few false positives, while a data-driven model could introduce some statistical classification errors. One engineer reported in an anecdotal case that one of his rule-based systems beat ML algorithms at detecting anomalous textual strings, because the patterns were so specific that if-else rules were perfectly effective compared to models trained with insufficient data. In general, however, as complexity grows, data-driven models scale better in performance than rules, which tend to require continuous exceptions to cover new cases, with inevitable qualitative degradation.

\subsubsection{System development and maintenance.} Creating a rules-based system requires an intensive initial knowledge engineering phase (collecting and formalizing rules from experts). Once implemented, maintenance consists of periodically checking the validity of the rules and updating them if the process changes; This can be laborious, but it is a process controlled by human experts. For a data-driven system, the initial phase is dedicated to data collection and preparation and model selection/training. Subsequent maintenance involves monitoring the performance of the model and retraining it with new data if necessary (to avoid drift). This can be burdensome if you don't have automated model update pipelines. In practice, the skills required to maintain the system differ: for the rules you still need the domain expert, for the ML model you need the data scientist. An often-overlooked aspect is that data-driven models can degrade invisibly – for example, if the distribution of data silently changes, the model may begin to give erroneous results without explicit warnings, until perhaps a fault is missed. On the other hand, a system of rules, even if antiquated, is difficult to break without someone noticing it (at most it does not cover a new case, but that case was explicitly outside its perimeter). This implies that data-driven systems require monitoring of monitoring, i.e. meta-systems that assess their health (e.g. alarms on models that begin to give out-of-distribution outputs).

\subsubsection{Cost and return on investment.} Rule-based systems can often be implemented with relatively low initial costs, especially if basic alarm schemes already exist in the control system (just configure thresholds and logics without additional hardware/software investments). The greatest costs are hidden in the man-time of the experts who define and update the rules. Data-driven systems may require investments in infrastructure (industrial databases, servers for analysis, ML software licenses) and staff training. On the other hand, once in operation, a good ML system can drastically reduce the downtime costs avoided thanks to early diagnostics, paying off the investment. The choice often depends on scale: in a small plant, one set of rules may be enough and cheap; in a smart factory with thousands of sensors, only an ML approach can fully exploit that wealth of data and prevent the many possible failure modes, justifying the investment.

\begin{figure}
\includegraphics[width=\textwidth]{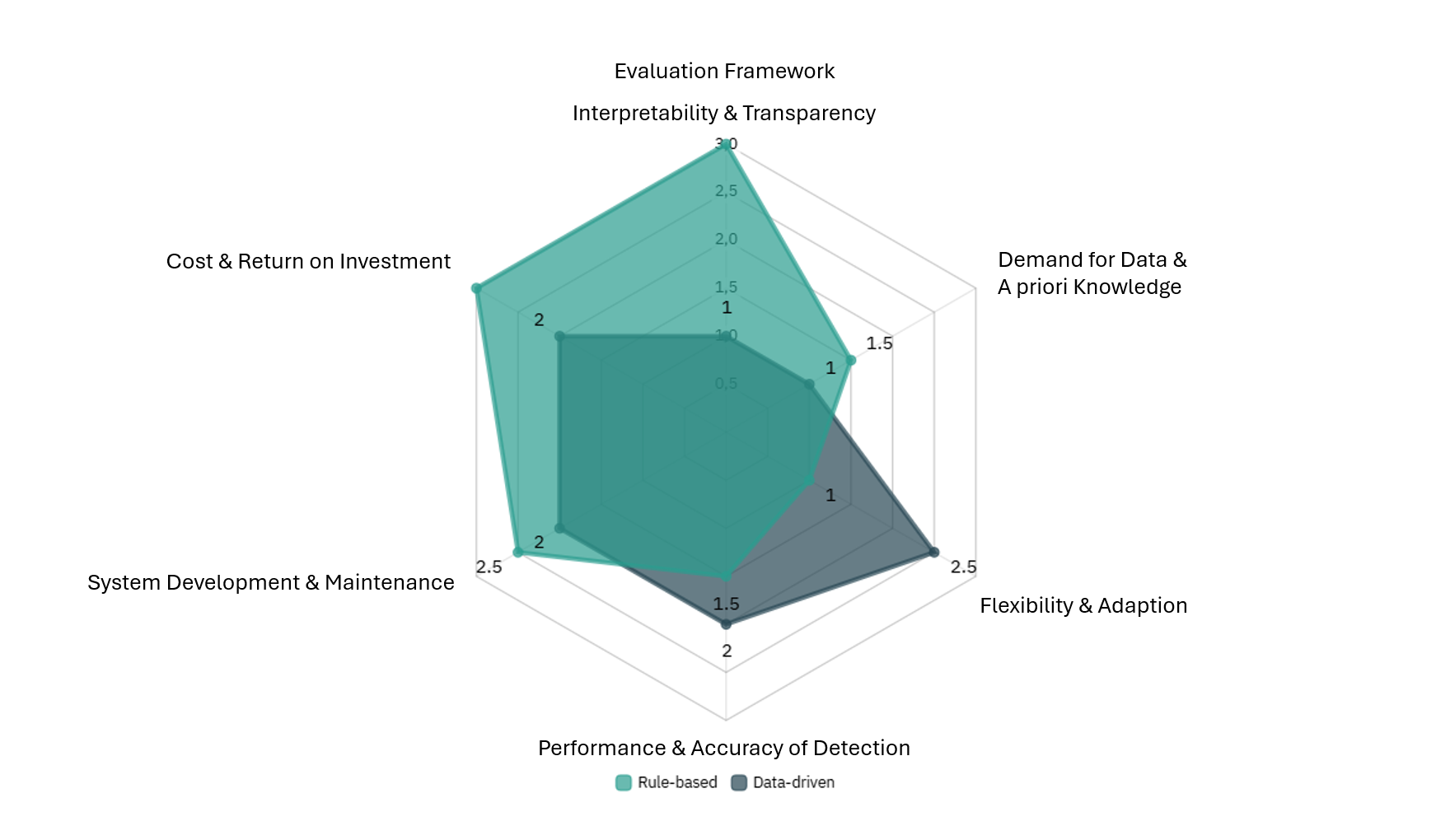}
\caption{The radar diagram represent the evaluation of Data-driven and Rule-based approaches considering six parameters on the axis with 3 possible different grades (0-low, 1-medium, 2-high) of performance.} \label{fig1}
\end{figure}

It should be noted that the comparison is not necessarily either or in many real-world applications, the two approaches coexist and can be integrated to achieve the best of both. For example, in the work of Nelson et al. on FDDs for buildings \cite{nelson2024machine} a hybrid framework is proposed where some simple rules filter basic situations or reduce the search space (preventing gross false alarms), while a more advanced ML module takes care of detecting complex patterns on the remaining residue. This synergy allows you to exploit the rules for their reliability in known conditions, and ML models to explore the unknown and refine the diagnosis. Even in manufacturing, hybrid approaches combine first-principles models (based on physical equations or control logics, similar to rules) with data-driven models, to constrain the statistical model within physically sensible limits, while improving its interpretability and robustness. Ultimately, direct comparison highlights complementary advantages: where one is lacking, the other excels. In Figure \ref{fig1} we propose a radar diagram to evaluate the six key features determined above to have a simple and quick comparison between the two approaches. Each feature is evaluated on a three values scale (low, medium, high performance); the system could be used to evaluate also hybrid solutions and compare them as one ore more feature changes.

\section{Discussion}\label{sec5}
From the above examination, the trade-offs between rule-based and data-driven systems clearly emerge, and it appears that the choice of the optimal approach strongly depends on the application scenario. Here are some key considerations and typical use cases.

\subsubsection{Ideal application scenarios for Rule-based approaches.} They find fertile ground in relatively stable contexts, where the phenomena to be monitored are well known. For example, in a mature manufacturing facility with processes that have been unchanged for years and abundant engineering knowledge accumulated, a rulebook system can be implemented quickly and easily monitored. Sectors with a strong safety culture or regulated (such as oil \& gas, aviation, pharmaceuticals) often prefer interpretable and certified solutions: a "mysterious" deep learning algorithm may not be acceptable to the authorities, while a set of deterministic logics can be verified and validated against regulatory standards. Even in applications where the event to be detected is binary and clearly defined (e.g., detecting if a robot has exceeded a torque limit, or if the temperature of a reactor is out of range), the threshold is a direct and easily fail-safe solution. Another favourable scenario is that of high-volume, low-variety lines: if the same part is always produced with few variations and the process is rigidly controlled, the failure modes are finite and known, so ad hoc rules can effectively cover them.
\subsubsection{Ideal application scenarios for Data-driven approaches.} They shine in situations where there is a great variety of conditions and complexities. For example, in the predictive maintenance of fleets of machinery (compressors, turbines, industrial vehicles) operating under different conditions, it is unthinkable to manually encode all the combinations of signals that indicate an incipient failure – much better to let an algorithm learn from data from vibrational, thermal, and electrical sensors. In the field of advanced Quality Control, for example in the visual inspection of products, classical rule-based computer vision systems (such as image filters and thresholds on colour or size of defects) have been largely supplanted by CNNs, which are able to recognize even very subtle or shape-varying defects, where static rules would fail \cite{mende2022}. For complex manufacturing processes (Industry 4.0), with hundreds of IoT sensors generating big data in real time, only data-driven methods can integrate all this information and provide useful analytics (detecting correlations between different machines, predicting bottlenecks, etc.). Finally, innovative sectors, such as additive manufacturing, industrial biotechnology, where there is a lack of established historical skills, benefit from learning algorithms that discover quality parameters and anomaly indicators independently.

\subsubsection{Current industrial trends. }
Although the academic literature in recent years has been dominated by proposals for data-driven methodologies and AI, in real industry the transition is gradual. Many companies have been running regulatory monitoring systems, sometimes for decades, because they are integrated into legacy control systems and trusted to be trusted. The 2024 PHM survey \cite{bouhadra2024knowledge} highlights that while research pushes towards data-driven techniques, companies still rely heavily on traditional knowledge-based methods. The reasons range from the fear of relying on black boxes, to the lack of internal skills, to the simplicity of what they know compared to the unknown of an ML algorithm. However, there is also an acceleration in the adoption of data-driven systems in the industrial sector, favoured by the spread of IIoT (Industrial IoT) platforms and more user-friendly industrial analytics software. Many vendors offer turnkey predictive maintenance solutions  where pre-trained models are applied to customer data, lowering the barrier to entry for companies that do not have in-house data scientists. Another driver is growing competitiveness: to remain efficient, enterprises seek to reduce unplanned downtime and improve quality – goals for which advanced data-driven systems have proven advantages.

\subsubsection{Open challenges and current research.}
Several lines of research aim to bridge the gap between the two approaches or mitigate the disadvantages highlighted. One is certainly Explainable AI applied to industry: making ML models more interpretable. For example, hybrid neuro-fuzzy systems or rule extraction from neural networks try to obtain models that learn from the data but return a set of understandable pseudo-rules \cite{monsef1997fuzzy}. Another strand is the integration of a priori knowledge into data-driven models: for example, physical or logical constraints can be incorporated into the cost function of a learning algorithm, so that the resulting model respects known properties (thus combining the benefits of a model based on first-principles physics with the flexibility of data-driven). This is particularly useful in the field of chemical and plant engineering processes, where there are conservation laws, mass/energy balances that must not be violated; purely data-driven models sometimes violate these principles by producing inconsistent results, which can be avoided by imposing constraints in the model \cite{musulin2013}. Transfer learning and few-shot learning approaches are also being studied  to address the problem of poor data: for example, using the trained model on a machine with a lot of data and transferring it to a similar one that has little fault data, so that we still have an initial data-driven system that is then perhaps refined with a few additional rules. At the same time, rule-based systems are also evolving we are talking about neuro-symbolic expert systems, where rules are supported by neural modules for parameters that are not easily defined, or rules automatically generated through historical data mining (e.g. extraction of if-then associations from data, transforming them into rules).

\subsubsection{Importance of the human factor. }
Regardless of the approach, it is crucial to involve domain experts in the development and use cycle of the monitoring system. In the case of rule-based systems, experts are the direct architects of the implemented logics, while in data-driven systems their role, although different, remains crucial: they must help select the right variables, validate the results of the model and make sure that the alarms "make sense" operationally. It was observed that the adoption of advanced monitoring systems often fails not because of technical deficiencies, but because of a lack of trust and usability on the part of field personnel. So, regardless of whether the core is rules-based or ML, the interface must present the information clearly, and perhaps allow the user to interact (e.g. by confirming or denying an alarm, providing feedback that in the data-driven case can be used to better re-train the model, and in the rule-based case to improve/recalibrate the rules). In short, artificial intelligence should be accompanied by human intelligence: the synergy between the two leads to the best results.

\section{Conclusions}\label{sec6}
In the industrial sector, the choice between a rules-based monitoring system and one based on data-driven approaches is not trivial and must be calibrated to specific needs, considering factors such as the availability of data, the need for interpretability, the complexity of the process and the resources available. Rule-based systems offer simplicity, transparency, and total control by human experts, making them ideal when you know your domain inside out and require deterministic reliability under known conditions. On the other hand, they suffer from covering unexpected scenarios and require manual updating in the face of changes. Data-driven systems, on the other hand, bring with them the power of machine learning: they can discover hidden anomalies and adapt to new situations, often with superior accuracy in recognizing failure patterns. However, they require abundant data and can appear opaque; By implementing them, continuous validation activities and the development of trust in the organization must be considered.
In terms of performance, the case studies highlight that properly trained data-driven methods tend to outperform rule-based systems in complex or highly variable contexts\cite{shiney} \cite{nelson2024machine}, helping to reduce downtime and production waste. Not surprisingly, many cutting-edge companies are investing in AI pilots for monitoring and predictive maintenance. That said, in stable and highly critical contexts, rule-based systems remain a reliable and often required reference for compliance (just think of nuclear warning systems, which are strongly based on deterministically verified logics).
In light of this analysis, a winning strategy seems to be not to see the two solutions as mutually exclusive, but to pursue a combined approach: using rules where they benefit (e.g. to ensure that values remain within safe limits, providing a simple and robust first line of defense) and flanking data-driven models for advanced monitoring which requires fine sensitivity and adaptation. AI models can be trained with existing knowledge (initializing their behavior to cover known cases right away) and then left free to learn more details from the data. In doing so, the strengths are combined: the reliability and clarity of the rules with the depth of analysis and flexibility of data driven.

In conclusion, the literature and experience converge in suggesting that the future of industrial monitoring will lie in intelligent hybrid solutions, where human knowledge and machine learning cooperate. This will allow you to achieve levels of performance and resilience that are unthinkable with traditional approaches alone, without losing the confidence and understanding necessary to operate in real environments. Such a synergy will enable truly smart industrial monitoring, in which systems not only react to problems, but anticipate and explain events, continuously improving thanks to feedback from the field and actively contributing to operational excellence.

\begin{credits}
\subsubsection{Declarations.} Authors occasionally used generative AI tools to improve the readability of some parts of the text. After using the tool, the authors reviewed and edited the content as needed and took full responsibility for the publication’s content.

\subsubsection{\discintname}
The authors have no competing interests to declare that are
relevant to the content of this article.
\end{credits}
%
%
%
%

\end{document}